# Cloud Adversarial Example Generation for Remote Sensing Image Classification

Fei Ma, *Member, IEEE*, Yuqiang Feng, Fan Zhang, *Senior Member, IEEE*
and Yongsheng Zhou, *Member, IEEE*

*Abstract*—Most existing adversarial attack methods for remote sensing images merely add adversarial perturbations or patches, resulting in unnatural modifications. Clouds are common atmospheric effects in remote sensing images. Generating clouds on these images can produce adversarial examples better aligning with human perception. In this paper, we propose a Perlin noise based cloud generation attack method. Common Perlin noise based cloud generation is a random, non-optimizable process, which cannot be directly used to attack the target models. We design a Perlin Gradient Generator Network (PGGN), which takes a gradient parameter vector as input and outputs the grids of Perlin noise gradient vectors at different scales. After a series of computations based on the gradient vectors, cloud masks at corresponding scales can be produced. These cloud masks are then weighted and summed depending on a mixing coefficient vector and a scaling factor to produce the final cloud masks. The gradient vector, coefficient vector and scaling factor are collectively represented as a cloud parameter vector, transforming the cloud generation into a black-box optimization problem. The Differential Evolution (DE) algorithm is employed to solve for the optimal solution of the cloud parameter vector, achieving a query-based black-box attack. Detailed experiments confirm that this method has strong attack capabilities and achieves high query efficiency. Additionally, we analyze the transferability of the generated adversarial examples and their robustness in adversarial defense scenarios.

## I. INTRODUCTION

WITH the advancement of satellite technology, high-resolution remote sensing images can now be obtained [1]. The abundance of high-quality data has significantly advanced research in various applications within the remote sensing field, such as scene classification [2], [3], semantic segmentation [4], and object detection [5]. Among these, remote sensing image classification is a crucial

This work was supported by the National Natural Science Foundation of China under Grant 62201027. (Corresponding author: Fan zhang.)

Fan Zhang is with the College of Information Science and Technology and the Interdisciplinary Research Center for Artificial Intelligence, Beijing University of Chemical Technology, Beijing 100029, China (e-mail: zhangf@mail.buct.edu.cn).

Fei Ma, Yuqiang Feng, and Yongsheng Zhou are with the College of Information Science and Technology, Beijing University of Chemical Technology, Beijing 100029, China (e-mail: mafei@mail.buct.edu.cn; yuqiangfg@gmail.com; zhyosh@mail.buct.edu.cn).

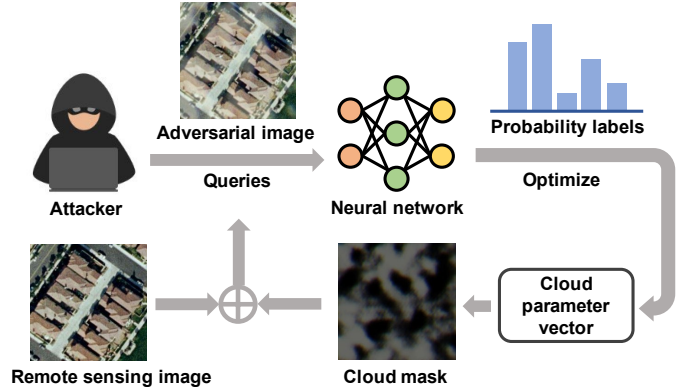

Fig. 1. The basic framework of the proposed method for generating cloud adversarial examples. The cloud parameter vector will be randomly initialized.

research area, finding applications in disaster detection [6], urban planning [7], and more. Currently, deep learning-based methods, particularly Convolutional Neural Networks (CNNs), dominate remote sensing image classification tasks due to their ability to extract robust and high-level features from images [8]. However, as deep learning models are extensively used to analyze remote sensing images, their vulnerability and uncertainty when exposed to adversarial examples require further investigation [9].

Szegedy et al. [10] first discovered that adding imperceptible perturbations to images could cause neural networks to output incorrect labels. These perturbed images are commonly known as "adversarial examples". Some adversarial attack methods, such as L-BFGS [10] and C&W [11], transform the generation of adversarial examples into a constrained optimization problem. Other attack methods, such as FGSM [12], BIM [13], and PGD [14], are gradient-based and aim to ascend the loss function gradient of the target models to deceive them. Some researchers also implement attacks by generating adversarial examples that appear realistic in the physical world. For instance, Wei et al. [15] used stickers in real life to deceive face classification neural networks by adjusting their placement and rotation angles. Zhong et al. [16] conducted attacks by simulating shadows on traffic signs. Liu et al. [17] generated adversarial examples by simulating raindrop images in nature.

The emergence of adversarial examples has also sparked interest among researchers in the remote sensing field. Some scholars began to investigate the feasibility of applying adversarial attack methods to remote sensing images. Czaja et al. [18] first implemented targeted attacks on remote sensing image classification neural networks. Additionally, studies have confirmed the existence of adversarial examples in Synthetic Aperture Radar (SAR) images and Hyperspectral Images (HSIs) [19], [20]. These studies indicate that remote sensing image classification neural networks, when faced with adversarial examples, will also misclassify with high confidence. Moreover, due to the relatively smaller amount of data in the remote sensing field compared to natural images [21], neural networks trained on these datasets may be more sensitive to adversarial examples.

Most adversarial attack methods in the current remote sensing field are white-box attacks. In a white-box scenario, the attacker has access to all information about the target models, including the network architectures, parameters, and even the training process. Given this detailed information, white-box attacks typically can generate more effective adversarial examples. Chen et al. [22] employed two representative gradient-based white-box attack methods, FGSM and BIM, to deceive various remote sensing image classification models. They found that the misclassification labels of adversarial examples exhibit an attack selectivity property, concentrating unevenly on specific categories. Xu et al. [23] measured the performance of FGSM on both targeted and untargeted attacks against remote sensing image classification models. They discussed the transferability of adversarial examples, indicating that adversarial examples obtained on a specific model may also mislead other models. Additionally, some attack methods employ adversarial patches with various scales or shapes to attack remote sensing object detection networks, causing them to misrecognize targets [24], [25], [26].

Unlike white-box attacks, black-box attacks simulate more realistic attack scenarios, where the attacker lacks internal knowledge of the target models. Currently, there are two main types of black-box attacks: transfer-based and query-based. Transfer-based attacks first compute perturbations on a surrogate model and then utilize the transferability between different models to effectively deceive the target models [27]. The primary goal of transfer-based methods is to enhance perturbation transferability. Xu et al. [28] observed that different remote sensing deep learning models might yield similar feature representations in the shallow layers. Based on this observation, they proposed a universal adversarial example generation method by attacking the shallow features of the surrogate model. Bai et al. [29] further proposed two variants of universal adversarial examples based on Xu's work [28]. The second type of black-box attacks are query-based attacks, which construct adversarial examples by querying the target models' output. Through multiple queries, the perturbations gain stronger attack capabilities while their intensity decreases, making them less perceptible to the human eyes. Among different query strategies, some methods attempt to directly estimate the gradients of the target models through numerical approximation [30]. Others use random search or evolutionary algorithms to make adversarial examples cross the decision boundary through iterations [31], [32].

Most of the above-mentioned remote sensing attack methods generate adversarial examples by adding adversarial perturbations or patches, but do not incorporate the uniqueness of remote sensing images relative to natural images. Atmospheric effects such as clouds and fog are common in remote sensing images, and these unique features can be utilized to generate adversarial examples. Inspired by this, some scholars have attempted to design relevant methods, but there are still many shortcomings. Tang et al. [33] designed natural weather-style perturbations to attack optical aerial detectors. This method simulates four weather phenomena on remote sensing images: snow, fog, shadows, and sun flares. Although the designed perturbations imitate the colors of these weather phenomena, they present unnatural ring-shaped or triangular morphologies. Sun et al. [34] proposed an adversarial cloud attack method for salient object detection in remote sensing images. They jointly tune adversarial exposure and additive perturbation for attack and constrain image close to random noise based cloudy image. The resulting adversarial examples contain unnatural noise, and the produced clouds entirely cover the original image. As a white-box attack method, it requires the gradient information of the target models, making it difficult to deploy in practice. From the perspective of application scenarios, the above two methods are limited to object detection and salient object detection. Currently, there is still a lack of relevant research on remote sensing image classification tasks.

To address the aforementioned issues, this paper proposes a black-box attack method to generate Perlin noise based cloud adversarial examples for remote sensing image classification. The basic framework of the proposed method is shown in Fig. 1. Perlin noise generates clouds similar in shape to those in real remote sensing images, without repetition, by randomly initializing parameters. In adversarial attack tasks, we need to control the shape, position, and color of the clouds so that the adversarial examples can deceive the target model. We design a Perlin Gradient Generator Network (PGGN). It takes the cloud parameter vector as input and outputs the grids of Perlin noise gradient vectors at various scales. These gradient vectors can subsequently generate a specific cloud mask. The cloud mask is then fused with the original remote sensing images to obtain the adversarial example. In this case, we can control the shape, position, and color of the cloud masks by altering the cloud parameter vector. The cloud parameter vector is updated based on the probability labels output by the target model. This process will be repeated until the generated cloud mask successfully deceive the target model.

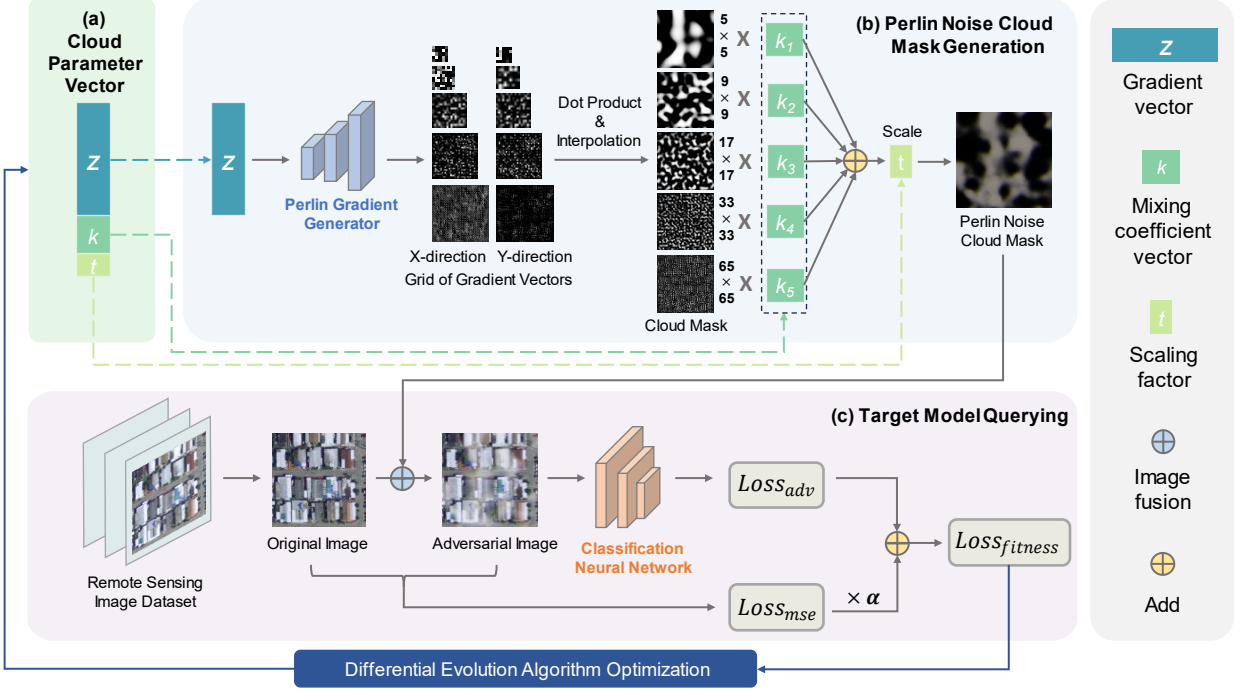

Fig. 2. An overview of the proposed method for cloud adversarial example generation with (a) cloud parameter vector, (b) Perlin noise cloud mask generation, (c) target model querying.

Compared to existing methods, the main contributions of the proposed method are as follows:

1) We propose a Perlin Gradient Generator Network (PGGN) to generate cloud masks. Traditionally, Perlin noise based cloud mask generation is a random, non-optimizable process. Our generator takes limited cloud parameters as input and outputs the grids of Perlin noise gradient vectors at different scales. A series of computations based on the gradient vectors can yield the corresponding cloud masks. By setting the cloud parameter vector, we can generate specific cloud masks while ensuring the natural appearance of clouds.

2) We propose a black-box attack method to generate cloud adversarial examples for remote sensing image classification. The generation of cloud adversarial examples is transformed into an optimization problem of the cloud parameter vector. The Differential Evolution (DE)[35] algorithm is employed to optimize the cloud parameter vector by repeatedly querying the target model's probability labels.

3) The attack capabilities, query efficiency and transferability of the proposed method are experimentally verified. Furthermore, we explore the attack selectivity of the adversarial examples and the potential impact of Perlin noise clouds on the neural networks.

The remainder of this paper is organized as follows. Section II introduces the proposed black-box attack method for generating cloud adversarial examples. Section III describes the experimental process and results. Section IV concludes the paper.

## II. METHODOLOGY

### A. Overview of the Proposed Method

Fig. 2 shows the overview of the proposed method for cloud adversarial example generation. The cloud parameter vector $r = [z, k, t]$ includes the gradient vector $z$, mixing coefficient vector $k$ and cloud thickness scaling factor $t$. First, the gradient vector $z$ is input into the trained PGGN to obtain five grids of gradient vectors at different scales. A series of calculations, such as dot products and interpolation on the grid's gradient vectors, yield cloud masks at corresponding scales. These cloud masks are then weighted and summed based on the mixing coefficient vector $k$ and scaling factor $t$ to produce the final Perlin noise cloud mask. Remote sensing images are combined with the cloud masks to generate the cloud adversarial examples. DE algorithm iteratively optimizes the cloud parameter vector based on the output of the target classification network until the number of queries reaches the upper limit or the attack is successful.

### B. Problem Description

Given an remote sensing image $I_{clear} \in \mathbb{R}^d$ and its corresponding label $y_{clear} \in [1, \cdots, m]$, a classification neural network $f : \mathbb{R}^d \to \mathbb{R}^m$ is trained to predict the label $\hat{y}_{clear}$ of input images:

$$\hat{y}_{clear} = \arg \max_i f_i(I_{clear}) \tag{1}$$

where $f_i(I_{clear})$ is the confidence of the $i-th$ class.

The proposed method generates a cloud adversarial example

$I_{adv}$ based on the original image $I_{clear}$, with optimization objective defined as follows:

$$\text{minimize } \mathcal{L}_{adv}\left(I_{adv}, y_{clear}\right) + \alpha * \mathcal{L}_{mse}\left(I_{adv}, I_{clear}\right)$$
$$s.t. \; y_{clear} \neq \hat{y}_{adv} \tag{2}$$

where $\mathcal{L}_{adv}$ is adversarial loss, which represents the confidence of class $y_{clear}$ when the adversarial example $I_{adv}$ is input into the target model. The mean square error loss $\mathcal{L}_{mse}$ measures the difference in pixel values between $I_{clear}$ and $I_{adv}$. $\alpha$ is the balance factor, a constant. The smaller the adversarial loss $\mathcal{L}_{adv}$, the more easily the generated adversarial examples are misclassified. Similarly, a lower mean square error loss $\mathcal{L}_{mse}$ indicates a smaller difference between the adversarial examples and the original images. The weighted sum of the two losses $\mathcal{L}_{adv}$ and $\mathcal{L}_{mse}$ is defined as the optimization objective to be minimized, causing the target model to misclassify the cloud adversarial examples while limiting the perturbations.

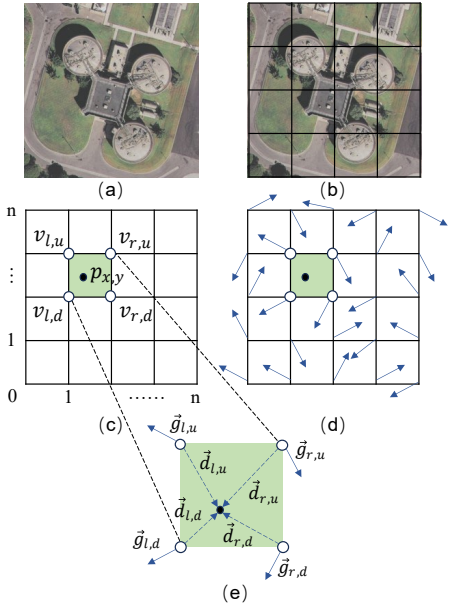

Fig. 3. Process of 2D Perlin noise generation. (a) Original remote sensing image (b) Grid division (c) $n \times n$ grid (d) Gradient vector assignment (e) Gradient vectors $\vec{g}$ and displacement vectors $\vec{d}$.

### C. Perlin Noise

Perlin noise [36], introduced by Ken Perlin in 1985, can be used to create highly realistic natural textures such as clouds, flames, and marble. It is a type of lattice gradient noise generated by interpolating or convolving random values (or gradients) defined at the points of the integer lattice [37]. Therefore, we chose to simulate clouds on remote sensing images based on Perlin noise. Perlin noise can be defined in any dimension, we only discuss the generation process of 2D Perlin noise here, which typically involves three main steps [38]:

**Step 1.** A $n \times n$ square grid is divided on the original remote sensing image (Fig. 3 (a)-(c)). Two-dimensional gradient vectors are assigned at each grid intersection, as shown in Fig. 3 (d).

**Step 2.** Given points $\mathcal{P} = \{p_{x,y} \mid 0 \leq x, y < n\}$, the four vertices of the small square where $p_{x,y}$ locates can be denoted as $\mathcal{V} = \{v_{i,j} \mid i = l, r; j = d, u\}$ where $l = \lfloor x \rfloor$, $r = l + 1$, $d = \lfloor y \rfloor$, $u = d + 1$. $\lfloor \bullet \rfloor$ represents the floor function.

As shown in Fig. 3 (e), the random gradient vectors at the four vertices are denoted as $\vec{g}_{i,j}$. The displacement vectors, which are the vectors from the vertices $v_{i,j}$ to $p_{x,y}$, can be expressed as:

$$\vec{d}_{i,j} = [x - i, y - j] \tag{3}$$

By computing the dot product of the gradient vectors and displacement vectors of each vertices, gradient ramp values $s_{i,j}$ can be obtained:

$$s_{i,j} = \vec{d}_{i,j} \cdot \vec{g}_{i,j} \tag{4}$$

**Step 3.** The interpolation function $fade(t) = 3t^2 - 2t^3$ is used on both two dimensions to blend the noise contributions from the four vertices. Let $\Delta x = x - l$ and $\Delta y = y - d$, a combined function $\sigma(\Delta x, \Delta y)$ can be defined as:

$$\sigma(\Delta x, \Delta y) = fade(\Delta x) \cdot fade(\Delta y) \tag{5}$$

The final Perlin noise value $perlin(x, y)$ at point $p_{x,y}$ is:

$$perlin(x, y) = \sigma(1 - \Delta x, 1 - \Delta y)s_{l,d} + \sigma(\Delta x, 1 - \Delta y)s_{r,d}$$
$$+ \sigma(1 - \Delta x, \Delta y)s_{l,u} + \sigma(\Delta x, \Delta y)s_{r,u} \tag{6}$$

For example, as point $p_{x,y}$ approaches the lower-left vertex $v_{l,d}$, $\lim_{x \to l}(1 - \Delta x) = 1$ and $\lim_{y \to d}(1 - \Delta y) = 1$. The value of the function $\sigma(1 - \Delta x, 1 - \Delta y)$ increases, indicating that the influence of the lower-left vertex on point $p_{x,y}$ becomes greater. This applies similarly to the other three vertices.

### D. Cloud Adversarial Examples Generation

For a square grid of size $n \times n$, $(n+1) \times (n+1)$ gradient vectors at grid vertices need to be initialized. The generation of Perlin noise is an uncontrollable process due to the randomness of the gradient vectors. The same gradient vectors, after a series of calculations such as dot product and interpolation, will result in the same cloud mask. Thus, controlling the clouds generated by Perlin noise can be simplified to controlling the grids of gradient vectors. To generate clouds with diverse shapes, the final cloud mask is obtained by the weighted summation of cloud masks corresponding to grids of different scales. This makes the total number of gradient vectors very large and it is hard to directly optimize each value.

To address this issue, we design PGGN to generate grids of gradient vectors at different scales, as shown in Fig. 4. The PGGN consists of a generator $\mathcal{G}$ and a discriminator $\mathcal{D}$. The generator $\mathcal{G}$ produces different grids of gradient vectors, while the discriminator $\mathcal{D}$ assesses the authenticity of the input grids.

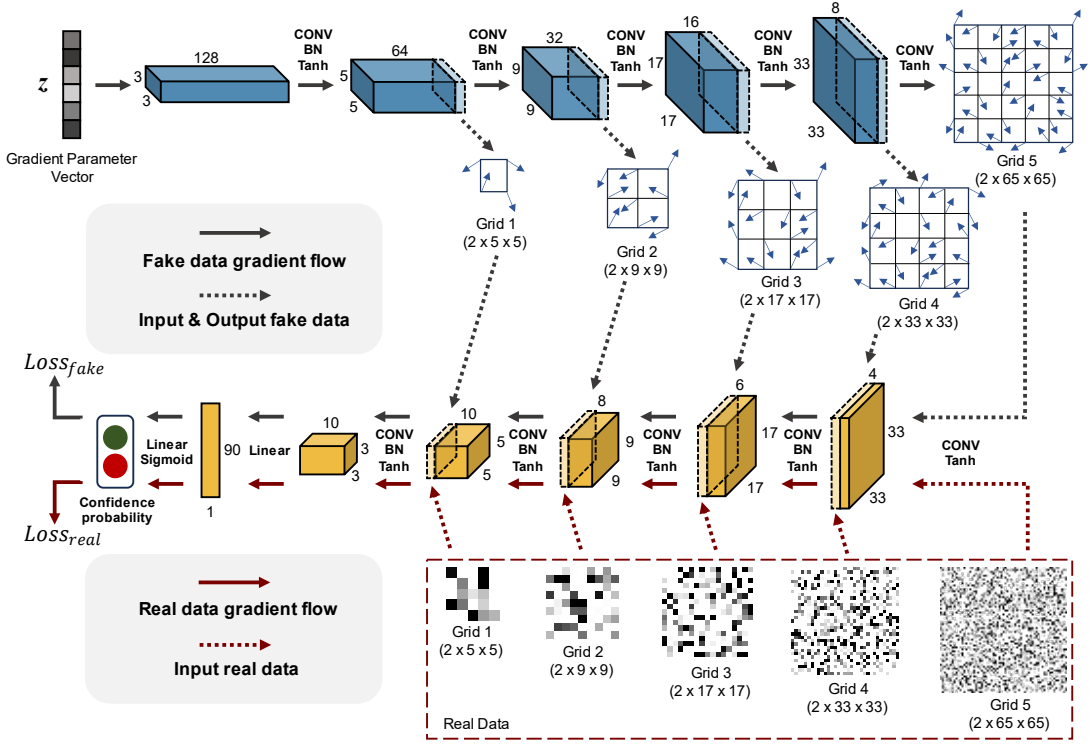

Fig. 4. Structure of the proposed Perlin Gradient Generator Network (PGGN). The upper blue section represents the generator $\mathcal{G}$, while the lower yellow section represents the discriminator $\mathcal{D}$.

In generator $\mathcal{G}$, the gradient parameter vector $z$ is transformed into a $3\times3$ feature with 128 channels through a fully connected layer, followed by multiple deconvolution operations that double the feature size while halving the number of channels. After the final deconvolution, the number of channels is reduced to two. The last two channels of each deconvolution output are selected as the grid of current size with gradient vectors. The two channels respectively represent the gradient vectors along x and y directions. The parameters of G are $\theta_g = \{\theta_g^1, \cdots, \theta_g^5\}$, where $\theta_g^i$ denotes the parameters required for generating the $i-th$ grid. Ultimately, the generator $\mathcal{G}$ maps $z$ to five grids of different sizes with gradient vectors $\mathcal{F} = \{\mathcal{G}(z; \theta_g^1), \cdots, \mathcal{G}(z; \theta_g^5)\}$. With the generator, specific cloud masks can be obtained by setting the gradient parameter vector, which greatly reduce the search difficulty of subsequent DE algorithm.

Given the discriminator $\mathcal{D}$ with parameters $\theta_d$, the largest grid $\mathcal{G}(z; \theta_g^5)$ is first input into the discriminator. After a convolution operation, the grid $\mathcal{G}(z; \theta_g^4)$ is concatenated into the feature map. Repeat the above process, reducing the feature size after each convolution, then concatenate the corresponding grid with gradient vectors.

The objective function of generator $\mathcal{G}$ can be expressed as:

$$\min_{\theta_g} \mathbb{E}_{z \sim p(z)} \log(1 - \mathcal{D}_{\theta_d}(\mathcal{F})) \tag{7}$$

The generator $\mathcal{G}$ aims to minimize objective such that $\mathcal{D}_{\theta_d}(\mathcal{F})$ is close to 1 (discriminator $\mathcal{D}$ is fooled into thinking generated $\mathcal{F}$ is real).

Let $\mathcal{R} = \{R_1, \cdots, R_5\}$ denote a set containing five real grids of gradient vectors. The objective function for training discriminator $\mathcal{D}$ can be expressed as:

$$\max_{\theta_d} \left[ \mathbb{E}_{\mathcal{R} \sim p_{data}} \log \mathcal{D}_{\theta_d}(\mathcal{R}) + \mathbb{E}_{z \sim p(z)} \log(1 - \mathcal{D}_{\theta_d}(\mathcal{F})) \right] \tag{8}$$

The discriminator $\mathcal{D}$ aims to maximize objective such that $\mathcal{D}_{\theta_d}(\mathcal{R})$ is close to 1 (real) and $\mathcal{D}_{\theta_d}(\mathcal{F})$ is close to 0 (fake).

After acquiring the grids with gradient vectors at five scales, the corresponding cloud masks $M_1, \cdots, M_5$ can be obtained through a series of calculations. These cloud masks of different sizes are weighed and summed according to the mixing coefficient vector $k = [k_1, \cdots, k_5]$ and normalized to $[0,1]$. In addition, the cloud thickness scaling factor $t$ is multiplied to adjust the values of the generated cloud. The final Perlin noise cloud mask $M$ is:

$$M = t \cdot \left( \frac{M_s - \min(M_s)}{\max(M_s) - \min(M_s)} \right) \tag{9}$$

where $M_s = \sum_{i=1}^{5} k_i \cdot M_i$, $\max(\bullet)$ and $\min(\bullet)$ represent taking the maximum value and the minimum value.

Czerkawski et al. [39] observed cloud characteristics in remote sensing images and proposed several modifications to Perlin noise, including Channel Misalignment and Channel-Specific Magnitude. We also incorporate these modifications to adjust the Perlin noise cloud masks, enhancing the realism of the generated clouds.

Inspired by previous work [40], [41], we fuse the Perlin noise cloud mask $M$ with the original remote sensing image

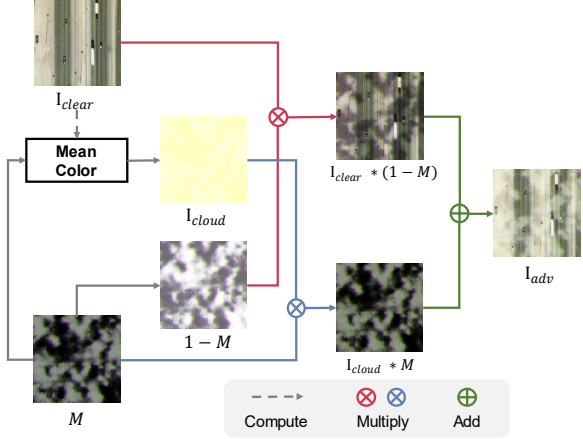

Fig. 5. The process of fusing Perlin noise cloud mask $M$ with original remote sensing image $I_{clear}$ to generate cloud adversarial example $I_{adv}$.

$I_{clear}$ to obtain the cloud adversarial example $I_{adv}$. As shown in Fig. 5, the mixing formula is:

$$I_{adv} = I_{clear} \cdot (1-M) + I_{cloud} \cdot M \quad (10)$$

The cloud component image $I_{cloud}$ is a mixed image of the cloud mask $M$ and the color mean of the cloud-free image $I_{clear}$.

### E. Cloud Adversarial Example Optimization

The DE algorithm is a population-based heuristic search method that can solve global optimization problems. It finds the optimal solution by simulating biological evolution processes such as mutation, crossover, and selection [35]. The simplicity and global search capability of the DE algorithm allow it to find the global optimal solution in multi-dimensional, multi-peak functions [42]. Unlike traditional optimization algorithms such as the gradient descent method, the DE algorithm does not require gradient information, enabling it to solve non-differentiable optimization problems. During the optimization of the Perlin noise cloud mask, the DE algorithm only needs to obtain the probability labels output by the target model, thereby achieving a query-based black-box attack.

The cloud parameter vector $r \in \mathbb{R}^q$ needs to be optimized is composed of gradient parameter vector $z$, mixing coefficient vector $k$ and cloud thickness scaling factor $t$. Population size of DE algorithm is denoted as $np$. $r_i(g)$ represents the $i-th$ individual (candidate solution) in the $g-th$ generation population, where $i = 1, 2, \cdots, np$. $r_{i,j}(g)$ represents the $j-th$ component of $r_i(g)$, where $j = 1, 2, \cdots, q$. The optimization process of DE algorithm is as follows:

**Step 1. Initialization:** At the initial stage $g = 0$, $np$ individuals $r_i(g)$ are randomly generated. Each component $r_{i,j}(g)$ of $r_i(g)$ is calculated by:

$$r_{i,j}(g) = r_{i,j}^L(g) + rand(0,1) \cdot (r_{i,j}^U(g) - r_{i,j}^L(g)) \quad (11)$$

where $r_{i,j}^U(g)$ and $r_{i,j}^L(g)$ represent the lower and upper bounds of the $r_{i,j}(g)$.

**Step 2. Mutation:** Three different individuals $r_{x_1}(g)$, $r_{x_2}(g)$, and $r_{x_3}(g)$ are randomly selected from the $g-th$ generation population, where $x_1 \neq x_2 \neq x_3$. The $i-th$ mutant individual $v_i(g+1)$ can be calculated as:

$$v_i(g+1) = r_{x_1}(g) + f \cdot (r_{x_2}(g) - r_{x_3}(g)) \quad (12)$$

where $f$ is the differential weight, a scalar.

**Step 3. Crossover:** The crossover operation is used to enhance the diversity of population. It selects components from the original individual $r_i(g)$ and mutant individual $v_i(g+1)$ to create new individual $u_i(g+1)$:

$$u_{i,j}(g+1) = \begin{cases} v_{i,j}(g+1), & \text{if } rand(0,1) \leq cr \\ r_{i,j}(g), & \text{otherwise} \end{cases} \quad (13)$$

where $cr$ is the crossover probability in the range 0 and 1.

**Step 4. Selection:** According to greedy strategy and the fitness function $\mathcal{L}_f$, select the better individual $r_i(g+1)$ from $u_i(g+1)$ and $r_i(g)$:

$$r_i(g+1) = \begin{cases} u_i(g+1), & \text{if } \mathcal{L}_f(u_i(g+1)) \leq \mathcal{L}_f(r_i(g)) \\ r_i(g), & \text{otherwise} \end{cases} \quad (14)$$

Using the cloud parameter vector $r$ as an example, the following explains how to calculate the fitness function $\mathcal{L}_f(r)$. Given the original remote sensing image $I_{clear}$ and the cloud parameter vector $r$, the cloud adversarial example $I_{adv}^r$ can be obtained according to section $D$. The cloud adversarial example $I_{adv}^r$ is input into the target model $f$, the confidence of the $i-th$ class can be denoted as $f_i(I_{adv}^r)$. If the correct label of the original image is $c$, for untargeted attacks, the adversarial loss can be defined as:

$$\mathcal{L}_{adv}(I_{adv}^r, c) = f_c(I_{adv}^r) \quad (15)$$

The smaller the adversarial loss $\mathcal{L}_{adv}$, the more likely the cloud adversarial example $I_{adv}^r$ is to mislead the target model. To limit the perturbation strength to the original images, we also calculate the mean square error (MSE) between the adversarial example $I_{adv}^r$ and the original image $I_{clear}$, termed the MSE loss:

$$\mathcal{L}_{mse}(I_{adv}^r, I_{clear}) = \frac{1}{p} \sum_{e=1}^{p} (I_{clear}(e) - I_{adv}^r(e))^2 \quad (16)$$

where $I(e)$ represents the $e-th$ pixel value of the image $I$, and $p$ is the total number of pixels in the image. The weighted sum of the adversarial loss $\mathcal{L}_{adv}$ and the MSE loss function allows DE algorithm to optimize the cloud adversarial example $I_{adv}^r$ while ensuring it remains as similar as possible to the original image $I_{clear}$.

$\mathcal{L}_{mse}$ is taken as the fitness function $\mathcal{L}_f(\mathbf{r})$ of the DE algorithm:

$$\mathcal{L}_f(\mathbf{r}) = \mathcal{L}_{adv}(I_{adv}^r, c) + \alpha \cdot \mathcal{L}_{mse}(I_{adv}^r, I_{clear}) \quad (17)$$

where $\alpha$ is the balance factor, a constant. This fitness

---

**Algorithm 1: Cloud Adversarial Example Generation**

**Input:** Cloud parameter vector $\mathbf{r} \in \mathbb{R}^q$, original image $I_{clear}$, target classifier $f$, correct label $c$, balance factor $\alpha$, number of population $np$, differential weight $f$, crossover probability $cr$, generation $g$, max queries $mq$.

**Output:** Cloud adversarial example: $I_{adv}$

1   $g \leftarrow 0$
2   $I_{adv} \leftarrow I_{clear}$
3   **for** $i = 1 \rightarrow np$ **do**
4     **for** $j = 1 \rightarrow q$ **do**
5       // Initialization with Eq.(11)
       $r_{i,j}(g) = r_{i,j}^L(g) + rand(0,1) \cdot (r_{i,j}^U(g) - r_{i,j}^L(g))$
6     **end**
7   **end**
8   **while** $(g*np < mq)$ **and** $(c = \arg\max_i f_i(I_{adv}))$ **do**
9     **for** $i = 1 \rightarrow np$ **do**
10      // Mutation with Eq.(12)
       $\mathbf{v}_i(g+1) = Mut(\mathbf{r}_{x_1}(g), \mathbf{r}_{x_2}(g), \mathbf{r}_{x_3}(g))$
11      **for** $j = 1 \rightarrow q$ **do**
12        // Crossover with Eq.(13)
        $u_{i,j}(g+1) = Cross(v_{i,j}(g+1), r_{i,j}(g))$
13      **end**
14      // Obtain the cloud adversarial examples
15      $I_{adv}^{r_i(g)} \Leftarrow \mathbf{r}_i(g); I_{adv}^{u_i(g+1)} \Leftarrow \mathbf{u}_i(g+1)$
16      // Compute fitness function with Eq.(15 - 17)
17      $\mathcal{L}_f(\mathbf{r}_i(g)) = \mathcal{L}_f(I_{adv}^{\mathbf{r}_i(g)}, I_{clear}, c, \alpha)$
18      $\mathcal{L}_f(\mathbf{u}_i(g+1)) = \mathcal{L}_f(I_{adv}^{\mathbf{u}_i(g+1)}, I_{clear}, c, \alpha)$
19      **if** $\mathcal{L}_f(\mathbf{u}_i(g+1)) \leq \mathcal{L}_f(\mathbf{r}_i(g))$ **then**
20        $\mathbf{r}_i(g+1) = \mathbf{u}_i(g+1)$
21        **if** $\mathcal{L}_f(\mathbf{r}_i(g+1)) \leq \mathcal{L}_f(\mathbf{r}_{best})$ **then**
22         $I_{adv} \leftarrow I_{adv}^{r_i(g+1)}$
23        **end**
24      **else**
25        $\mathbf{r}_i(g+1) = \mathbf{r}_i(g);$
26      **end**
27     **end**
28     $g \leftarrow g+1;$
29   **end**
30   **return** $I_{adv}$

---

The algorithm iterates steps 2-4. When the number of queries (the number of iterations multiplied by the population size $np$) reaches the set limit or the generated adversarial example is misclassified by the target model, the algorithm terminates. The overall algorithm flow of the proposed method is shown in Algorithm 1.

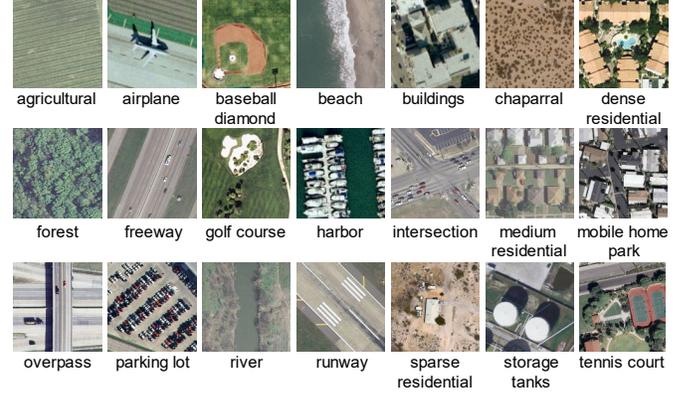

Fig. 6. Example images of each category in the UCM dataset, with the category labeled below each image.

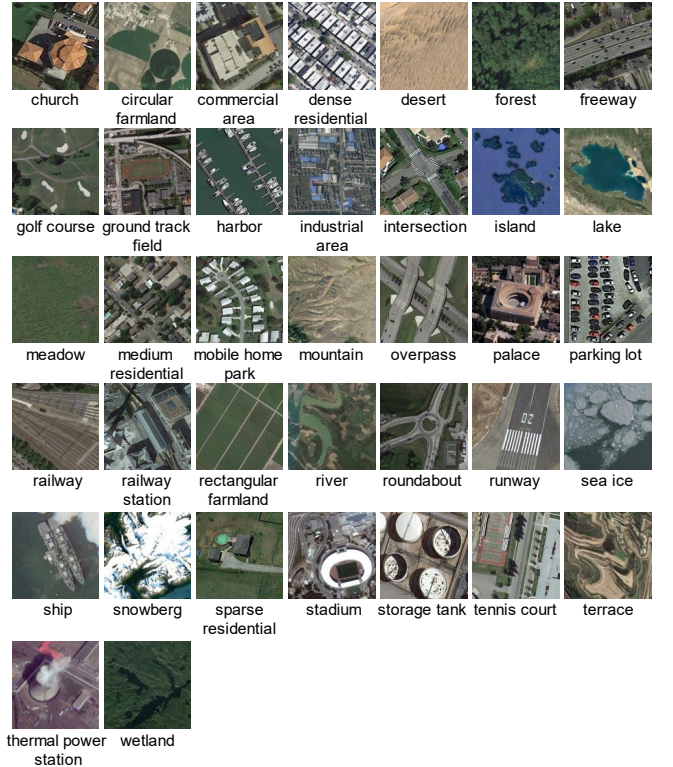

Fig. 7. Example images of each category in the NWPU dataset, with the category labeled below each image.

## III. EXPERIMENTS

To verify the effectiveness of the proposed method, we conduct multiple untargeted attack experiments on two remote sensing image classification datasets: UC Merced Land Use [43] and NWPU-RESISC45 [21]. Section A introduces the datasets, target models, comparison methods, and evaluation metrics. Section B explains the experimental settings of the proposed method. Section C presents the quantitative results, visualization, adversarial example label distribution, and parameter analysis. Section D explains the potential impact of Perlin noise clouds on classification networks. Section E measures the transferability of the generated adversarial examples across different models. Section F further studies the robustness of the generated adversarial examples.

TABLE I
TEST ACCURACY (%) OF DIFFERENT MODELS ON UCM AND NWPU DATASETS

| Dataset | AlexNet | VGG16 | ResNet18 | ResNet101 | DenseNet121 | DenseNet201 | Inception-v3 | RegNetX-400MF |
|---------|---------|-------|----------|-----------|-------------|-------------|--------------|----------------|
| UCM | 90.71 | 94.29 | 90.00 | 93.57 | 92.38 | 94.29 | 91.90 | 90.95 |
| NWPU | 86.69 | 91.53 | 90.24 | 94.43 | 93.47 | 94.06 | 91.49 | 91.01 |

TABLE II
PARAMETER SETTINGS OF DIFFERENT COMPARISON METHODS

| Attack Method | Parameter | Distance Measure |
|---------------|-----------|------------------|
| FGSM [12] | $eps = 0.05$ | $L_\infty$ |
| BIM [13] | $eps = 0.05, alpha = eps / 10$<br>$steps = 5$ | $L_\infty$ |
| DeepFool [51] | $overshoot = 0.02, steps = 50$ | $L_2$ |
| FAB [52] | $eps = 0.05, steps = 10$<br>$n\_restarts = 1, eta = 1.05$<br>$alpha\_max = 0.1, beta = 0.9$ | $L_\infty$ |
| C&W [11] | $c = 1, \kappa = 0, \eta = 0.01$<br>$steps = 50$ | $L_2$ |
| SimBA-DCT [31] | $overshoot = 0.5, freq\_dim = 4$<br>$stride = 1, max\_iter = 3000$ | $L_2$ |
| Square-Attack [53] | $eps = 0.05, n\_restarts = 1$<br>$n\_queries = 3000$ | $L_\infty$ |

## A. Datasets and Comparison Methods

*1) Datasets:* The UC Merced Land Use (UCM) dataset is a 21-class land use image dataset with 100 images per class, each image sized at 256×256 pixels. These images are manually extracted from the US Geological Survey (USGS) National Map Urban Area Imagery Collection, covering various urban areas across the country, with pixel resolution of 1 foot. The sample images are shown in Fig. 6.

The NWPU-RESISC45 (NWPU) dataset is a publicly available benchmark dataset for remote sensing image scene classification. This dataset contains 31,500 images covering 45 scene classes, with 700 images per class. Since the dataset includes a cloud category, and our method aims to simulate clouds to generate adversarial examples, we exclude this category to avoid potential impact. Therefore, in this experiment, we only use the remaining 44 categories, totaling 30,800 images, as shown in Fig. 7.

*2) Target Models:* For the two datasets, we randomly select 80% as the training set and the remaining 20% as the test set. We choose commonly used models in remote sensing image classification tasks, including AlexNet [44], VGG16 [45], ResNet18 [46], ResNet101 [46], DenseNet121 [47], DenseNet201 [47], Inception-v3 [48], and RegNetX-400MF [49], as target models. The experimental platform is based on Ubuntu 18.04.6 LTS and the PyTorch framework [50], with AMD EPYC 7551P 32-Core Processor CPU and NVIDIA RTX A5000 GPU. Each model is trained separately on the UCM and NWPU datasets based on pre-trained weights. The batch size is 64, and the learning rate is 1e-4. The momentum factor of the SGD optimizer is 0.9, and the weight decay is 5e-

4. The input size of the images is 256×256, and data augmentation operations, such as random vertical or horizontal flips are used during training. The test accuracy of each model on the two datasets is shown in Table I.

TABLE III
PARAMETER SETTINGS OF THE PROPOSED METHOD

| Parameter | Value |
|-----------|-------|
| $q(z \in \mathbb{R}^q)$ | 52 |
| $z^L$ | [-1, -1, …, -1] |
| $z^U$ | [1, 1, …, 1] |
| $k^L$ | [0, 0, 0, 0.4, 0.6] |
| $k^U$ | [0.1, 0.2, 0.3, 0.8] |
| $t^L$ | 0.1 |
| $t^U$ | 0.65 |
| $np$ | 100 |
| $cr$ | 0.80 |
| $f$ | 0.50 |
| $\alpha$ | 0.25 |
| $mq$ | 3000 |

*3) Comparison Methods:* We choose five white-box attack methods: FGSM [12], BIM [13], DeepFool [51], FAB [52], C&W [11], and two query-based black-box attack methods: SimBA-DCT [31] and Square Attack [53], to compare with our proposed method. The specific parameter settings for each attack method are shown in Table II, such as maximum perturbation (epsilon), step size (alpha), learning rate, confidence, number of algorithm iterations (steps), max queries, distance measure, etc.

*4) Metrics:* To measure the effectiveness of the attack methods, we use Attack Success Rate (ASR, %) as the evaluation metric, which can be defined as:

$$\text{ASR} = \frac{n_{adv}}{n_{total} - n_{misclassified}} \tag{18}$$

where $n_{total}$ represents the total number of test images. $n_{misclassified}$ represents the number of images that are already misclassified before attack, and thus there is no need to generate adversarial examples for them. $n_{adv}$ is the number of adversarial examples successfully misleading the target models.

For query-based attacks, we additionally use Average Queries (AQ) to compare the optimization complexity:

$$AQ = \frac{\sum_{i=1}^{n_{adv}} w^i}{n_{adv}} \quad (19)$$

where $w^i$ represents the number of queries needed for the $i-th$ adversarial example. AQ is a crucial metric for evaluating the efficiency of query-based attacks. Lower AQ indicates higher attack efficiency.

TABLE IV
ATTACK SUCCESS RATE (ASR, %) OF DIFFERENT METHODS ON UCM DATASET

| Type | Attack method | AlexNet | VGG16 | ResNet18 | ResNet101 | DenseNet121 | DenseNet201 | Inception-v3 | RegNetX-400MF | Average |
|------|--------------|---------|-------|----------|-----------|-------------|-------------|--------------|---------------|---------|
| White box attack | FGSM [12] | 96.14 | 91.56 | 76.88 | 58.13 | 66.94 | 63.32 | 47.01 | 58.84 | 69.85 |
| | BIM [13] | 99.45 | 99.20 | 99.72 | 97.86 | 98.92 | 99.20 | 99.73 | 98.61 | **99.09** |
| | DeepFool [51] | 93.09 | 86.70 | 71.03 | 68.98 | 86.18 | 83.86 | 80.65 | 75.41 | 78.97 |
| | FAB [52] | 88.67 | 95.23 | 95.81 | 91.73 | 85.60 | 88.56 | 88.86 | 85.75 | 90.03 |
| | C&W [11] | 92.62 | 98.68 | 98.61 | 92.76 | 96.75 | 96.28 | 99.18 | 98.62 | 96.69 |
| Query-based black box attack | SimBA-DCT [31] | 97.52 | 87.53 | 96.38 | 87.67 | 91.87 | 87.53 | 85.87 | 84.30 | 89.83 |
| | Square-Attack [53] | 95.32 | 97.61 | 99.72 | 99.20 | 99.73 | 99.47 | 99.73 | 99.17 | 98.74 |
| | **Proposed** | 93.70 | 87.12 | 92.86 | 84.99 | 88.14 | 89.39 | 94.82 | 94.74 | 90.72 |

TABLE V
ATTACK SUCCESS RATE (ASR, %) OF DIFFERENT METHODS ON NWPU DATASET

| Type | Attack method | AlexNet | VGG16 | ResNet18 | ResNet101 | DenseNet121 | DenseNet201 | Inception-v3 | RegNetX-400MF | Average |
|------|--------------|---------|-------|----------|-----------|-------------|-------------|--------------|---------------|---------|
| White box attack | FGSM [12] | 98.59 | 97.19 | 89.50 | 79.15 | 81.68 | 70.51 | 61.27 | 77.58 | 81.93 |
| | BIM [13] | 100.00 | 100.00 | 100.00 | 99.79 | 99.36 | 97.20 | 99.78 | 99.79 | 99.49 |
| | DeepFool [51] | 93.45 | 85.32 | 63.30 | 69.60 | 85.81 | 88.19 | 85.90 | 75.82 | 80.92 |
| | FAB [52] | 84.35 | 96.75 | 94.48 | 94.12 | 93.29 | 92.98 | 88.02 | 86.84 | 91.35 |
| | C&W [11] | 99.77 | 99.78 | 100.00 | 99.37 | 98.92 | 98.10 | 99.36 | 99.56 | 99.36 |
| Query-based black box attack | SimBA-DCT [31] | 98.19 | 91.80 | 99.34 | 93.86 | 94.84 | 88.17 | 89.32 | 94.00 | 93.69 |
| | Square-Attack [53] | 99.76 | 99.77 | 99.55 | 99.79 | 99.58 | 100.00 | 100.00 | 100.00 | **99.81** |
| | **Proposed** | 97.40 | 95.00 | 94.60 | 91.40 | 95.40 | 91.80 | 96.60 | 95.60 | 94.73 |

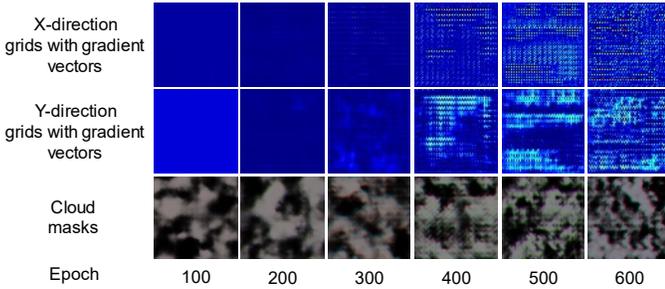

Fig. 8. Grids with gradient vectors generated at different training stages and the corresponding cloud masks. From top to bottom: X-direction 65×65 grids with gradient vectors, Y-direction 65×65 grids with gradient vectors, corresponding cloud masks, and training epoch.

### B. Experimental Settings of the Proposed Method

During training of PGGN, the five sizes of grids are 4×4, 8×8, 16×16, 32×32, and 64×64 pixels. For each size, a batch of 5000 grids with gradient vectors is generated using the traditional Perlin noise method as training data. The Adam optimizer's learning rate is 2e-4, with the first-order momentum decay factor $\beta_1$ =0.5, and the second-order momentum decay factor $\beta_2$ =0.999. The number of training epochs is 600, and the gradient vector's dimension $q$ is 52. Fig. 8 visualizes the grids with gradient vectors generated at different training stages and the corresponding Perlin noise cloud masks.

During the attack, the cloud masks are used to obtain the adversarial examples through image fusion. The differences between the adversarial examples and the original images are the directly applied perturbations, not the cloud masks. Adversarial examples and their corresponding cloud masks and perturbations are illustrated in Fig. 9.

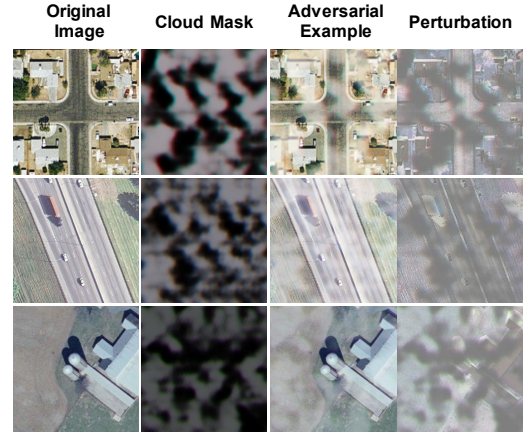

Fig. 9. Adversarial examples and corresponding cloud masks and perturbations.

The parameters of DE algorithm are detailed in Table III. These include the upper and lower search bounds of variable, i.e., $z^L$, $z^U$, $k^L$, $k^U$, $t^L$, $t^U$, the population size $np$, the crossover probability $cr$, and the differential weight $f$. $\alpha$ is set to 0.25 to balance the attack strength $\mathcal{L}_{adv}$ and the perturbation effect $\mathcal{L}_{mse}$. The max queries of the algorithm $mq$ is consistent with other query-based attack methods.

### C. Experimental Results and Analysis

TABLE VI
AVERAGE QUERIES (AQ) OF DIFFERENT QUERY-BASED BLACK-BOX ATTACK METHODS ON UCM DATASET

| Distance measure | Attack method | AlexNet | VGG16 | ResNet18 | ResNet101 | DenseNet121 | DenseNet201 | Inception -v3 | RegNetX -400MF | Average |
|---|---|---|---|---|---|---|---|---|---|---|
| $L_2$ | SimBA-DCT [31] | 737 | 740 | 427 | 601 | 500 | 610 | 551 | 479 | 581 |
| $L_\infty$ | Square-Attack [53] | 419 | 354 | 181 | 206 | 189 | 208 | 156 | 120 | 229 |
| - | **Proposed** | 199 | 210 | 220 | 226 | 213 | 208 | 185 | 196 | **207** |

TABLE VII
AVERAGE QUERIES (AQ) OF DIFFERENT QUERY-BASED BLACK-BOX ATTACK METHODS ON NWPU DATASET

| Distance measure | Attack method | AlexNet | VGG16 | ResNet18 | ResNet101 | DenseNet121 | DenseNet201 | Inception -v3 | RegNetX -400MF | Average |
|---|---|---|---|---|---|---|---|---|---|---|
| $L_2$ | SimBA-DCT [31] | 356 | 445 | 391 | 445 | 466 | 516 | 436 | 413 | 434 |
| $L_\infty$ | Square-Attack [53] | 106 | 93 | 68 | 106 | 135 | 142 | 96 | 36 | **98** |
| - | **Proposed** | 125 | 115 | 139 | 178 | 177 | 136 | 157 | 165 | 149 |

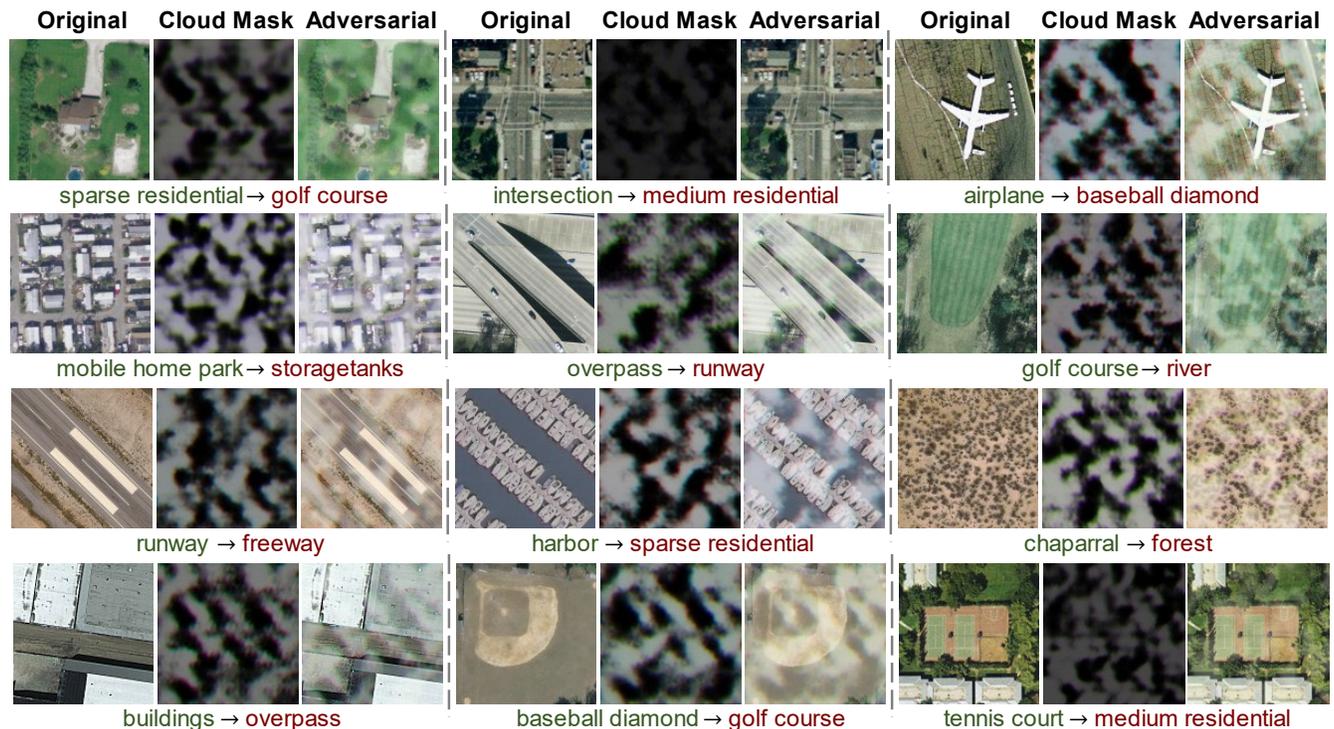

Fig. 10. Adversarial examples generated by the proposed method on UCM dataset. Each group of images, from left to right, the original images, the Perlin noise cloud masks, and the corresponding adversarial examples. Green labels indicate the correct categories and red labels indicate the labels predicted by the target models.

*1) Quantitative Analysis:* To evaluate the effectiveness of adversarial attack methods, 400 images were randomly selected from the test set of the UCM dataset and 500 images from the test set of the NWPU dataset. Tables IV-V present the untargeted ASRs of different methods. Our proposed method achieves an average ASR exceeding 90% on both datasets, demonstrating its effective attack capability. Our method's ASR surpasses that of FGSM, DeepFool, and FAB, but falls short of BIM and C&W when compared with white-box attack methods. In comparison with black-box attack methods, our method outperforms SimBA-DCT but is less effective than Square Attack.

Tables VI-VII compare the AQ of three query-based attacks on the UCM and NWPU datasets. Our proposed method demonstrates high query efficiency. For instance, on the UCM dataset, the AQ of our method across different target models is 207, which is lower than both SimBA-DCT and Square Attack. On the NWPU dataset, the AQ of our method is 148, slightly

higher than Square Attack's 98 but significantly lower than SimBA-DCT's 434.

Adversarial examples generated by the proposed method on the UCM and NWPU datasets are shown in Figs. 10-11. Our method simulates the shape and color of clouds in remote sensing images, producing adversarial examples that are more visually deceptive than those generated by typical adversarial attack methods.

Note that in the experiments described above, we do not directly constrain the perturbations of the proposed method under $L_2$ or $L_\infty$ norms. Most adversarial attack methods aim to achieve high ASRs with minimal perturbation norms. However, our method focuses on generating natural cloud adversarial examples. Under these circumstances, the $L_2$ or $L_\infty$ norms of the perturbations might not be reasonable metrics. For instance, in Fig. 12, we generate adversarial examples for the same remote sensing images using our method, as well as SimBA-DCT and Square Attack for comparison. Although the perturbations of our method are larger under $L_2$ or $L_\infty$ norms, they appear more natural from a human visual perspective.

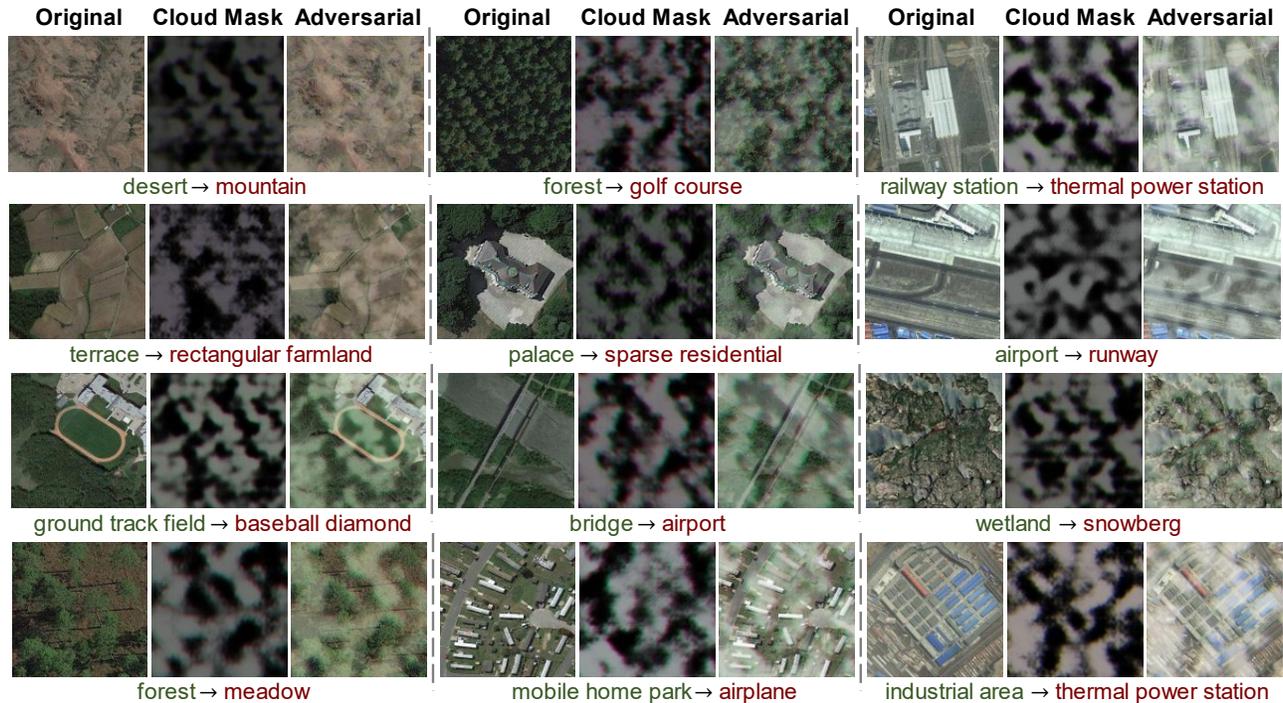

Fig. 11. Adversarial examples generated by the proposed method on NWPU dataset. Each group of images, from left to right, the original images, the Perlin noise cloud masks, and the corresponding adversarial examples. Green labels indicate the correct categories and red labels indicate the labels predicted by the target models.

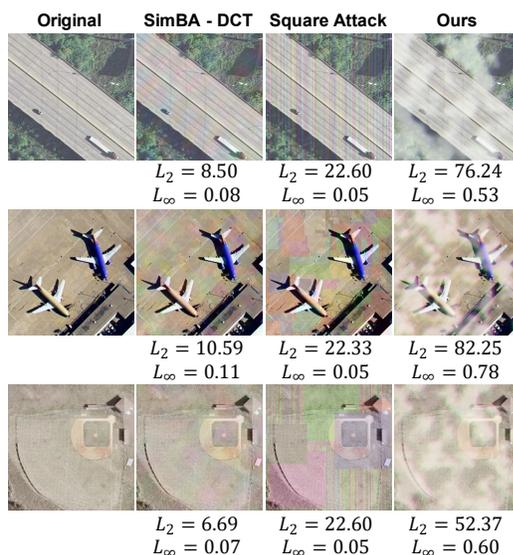

Fig. 12. Adversarial examples generated by three different methods. From left to right: the original images, the adversarial examples generated by SimBA-DCT, Square Attack, and proposed method. The $L_2$ and $L_\infty$ norms of the perturbations are annotated below each adversarial example.

We analysis the perturbations from both shape and color perspectives. SimBA-DCT's adversarial perturbations are generally smoother in shape, but the applied color is still noticeable. Square Attack uses colorful block and vertical line perturbations, resulting in adversarial examples with obvious unnatural textures. Our method simulates the natural shape of clouds and retains the original image colors. Additionally, clouds are a natural atmospheric phenomenon in remote sensing images, making cloud adversarial examples more plausible and aligned with human cognition. Therefore, our adversarial examples are less likely to be detected by the human eye compared to other query-based black-box methods

*2) Attack Selectivity:* Previously, Chen et al. [54] discussed attack selectivity in the context of remote sensing image classification, indicating that the categories of generated adversarial examples are not randomly and uniformly distributed but are selectively concentrated in certain categories. As shown in Fig. 13, we also analyze the classification confusion matrix of our adversarial examples on the UCM dataset. The column and row labels represent the correct labels of the original images and predicted labels of the generated adversarial examples. It can be observed that attack selectivity is related to the target neural networks' structures. For example, VGG16 tends to classify adversarial examples as category 11 (Intersection), while DenseNet121 tends to classify them as category 18 (Sparse residential). This may be due to differences in the classification decision boundaries of neural networks, causing images to move more easily into specific classification regions when perturbed. Additionally, the inherent characteristics of the images also affect attack selectivity. Regardless of the target model, images with the category 2 (Baseball diamond) are mostly classified as category 9 (Golf course) when perturbed. From the perspective of image morphology, baseball fields and golf courses are quite similar, suggesting that their corresponding high-dimensional feature vectors are close in the target models .

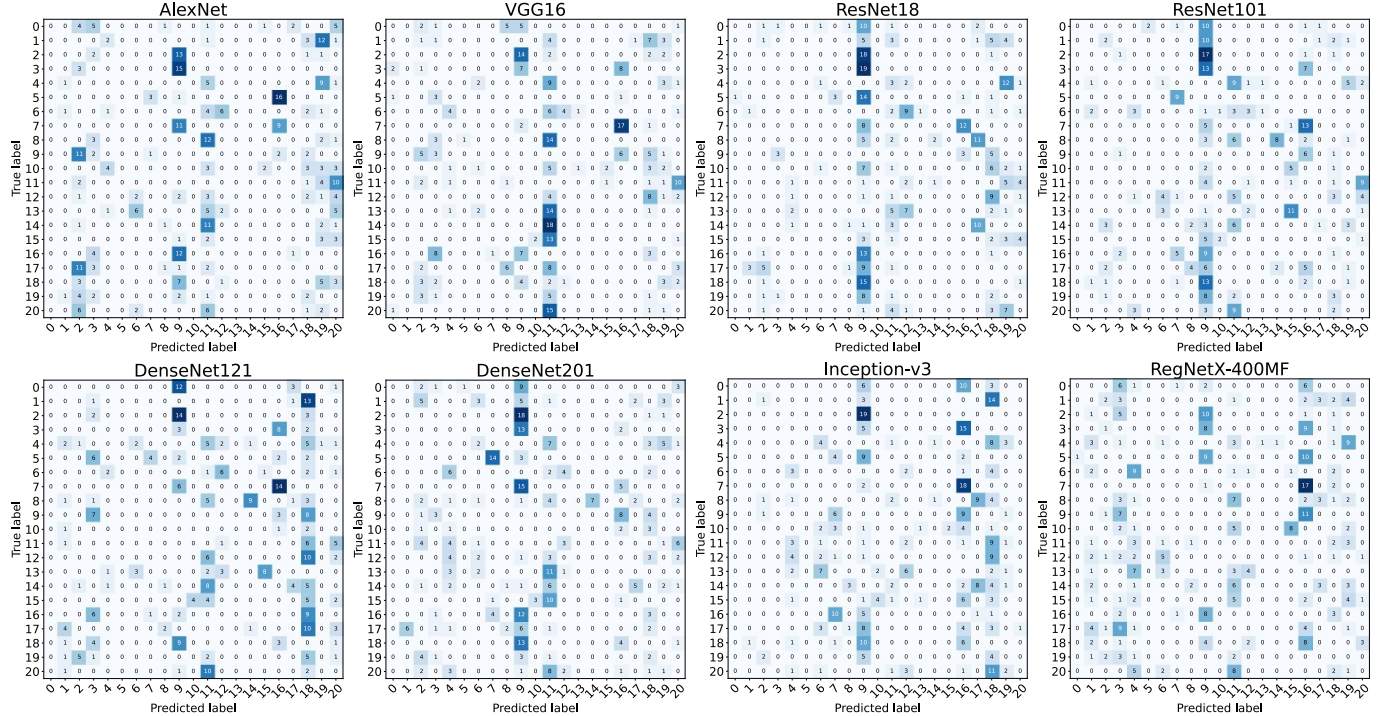

Fig. 13. Confusion matrix of different target models for cloud adversarial examples on the UCM dataset. The y-axis 'True label' represents the correct labels of adversarial examples, and the x-axis 'Predicted label' represents the predicted labels by target models. The label numbers from 0 to 20 represent the categories as follows: Agricultural, Airplane, Baseball diamond, Beach, Buildings, Chaparral, Dense residential, Forest, Freeway, Golf course, Harbor, Intersection, Medium residential, Mobile home park, Overpass, Parking lot, River, Runway, Sparse residential, Storage tanks, Tennis court.

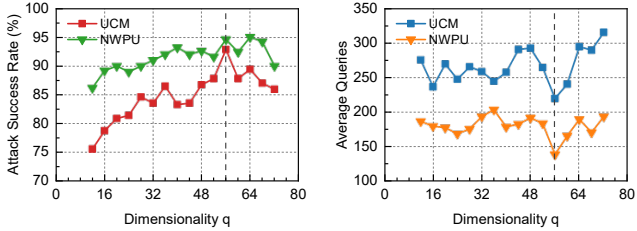

Fig. 14. ASR (left) and AQ (right) of the proposed method for different dimensionality $q$ on the UCM and NWPU datasets.

Therefore, these images are easily misclassified into other morphologically similar categories.

*3) Parameter Analysis:* The dimensionality $q$ of the gradient parameter vector $z$ in the PGGN affects the cloud generation and the attack efficiency. We train the PGGN using different dimensionalities $q$ and attack the ResNet18 on both UCM and NWPU datasets. The resulting ASR and AQ are shown in Fig. 14. As the dimensionality $q$ increases from 16 to 80 with an interval of 4, the ASR initially increases and then slightly decreases, while the AQ does not show a clear trend. It is speculated that when the dimensionality $q$ is small, the trained generator cannot effectively control the Perlin noise clouds, resulting in a lower ASR. Conversely, when the dimensionality $q$ is too large, the search space for the DE algorithm also expands, increasing the search difficulty and negatively impacting the generation of the Perlin noise airplane    tennis court    baseball diamond baseball diamond    intersection    golf course clouds. When $q = 56$, the proposed attack method achieves the highest overall ASR and the lowest AQ on both datasets.

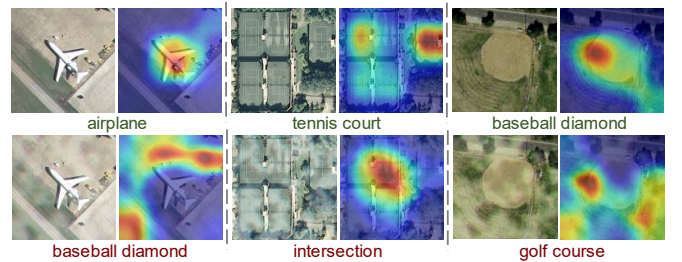

Fig. 15. CAMs of ResNet18. From top to bottom, original images with their CAMs, adversarial examples with their CAMs. The green and red labels indicate the true categories and predicted results by the target models.

### D. Why Perlin Noise Clouds Work

To explore why Perlin noise clouds mislead neural networks, we use Grad-CAM [55] to visualize the class activation maps (CAMs) of ResNet18 for remote sensing images before and after adding Perlin noise cloud perturbations. These maps indicate which parts of the input image significantly impact the neural network's outputs. As shown in Fig. 15, after adding Perlin noise clouds, there is a notable change in the areas of the images that the neural network attends to. For instance, in the leftmost image, the neural network initially focuses on the airplane, which is the central object. After adding perturbations, its attention completely shifts to the surrounding background. Visually, the structures of objects in the images are not significantly damaged, demonstrating the effectiveness of the generated Perlin noise clouds.

TABLE VIII
TRANSFER ATTACK SUCCESS RATE (TASR) OF THE PROPOSED METHOD ON UCM DATASET

| Surrogate Model | Target Model | | | | | | | | |
| | AlexNet | VGG16 | ResNet18 | ResNet101 | DenseNet121 | DenseNet201 | Inception-v3 | RegNetX-400MF | **Average** |
|---|---|---|---|---|---|---|---|---|---|
| AlexNet | - | 33.89 | 43.42 | 32.21 | 38.94 | 29.41 | 44.54 | 40.90 | 37.62 |
| VGG16 | **52.17** | - | 51.59 | 37.97 | 43.77 | 31.88 | 46.96 | 45.80 | 44.31 |
| ResNet18 | 46.44 | 31.91 | - | 29.63 | 40.74 | 31.91 | 46.72 | 38.75 | 38.01 |
| ResNet101 | 47.60 | **41.02** | **55.69** | - | **48.80** | **40.42** | 54.19 | **47.31** | **47.86** |
| DenseNet121 | 44.44 | 35.67 | 48.83 | 36.55 | - | 35.96 | 47.95 | 47.08 | 42.35 |
| DenseNet201 | 46.89 | 38.98 | 48.59 | **38.70** | 44.92 | - | 50.56 | 43.22 | 44.55 |
| Inception-v3 | 41.53 | 30.05 | 42.90 | 27.05 | 37.43 | 26.78 | - | 42.62 | 35.48 |
| RegNetX-400MF | 40.00 | 30.28 | 39.44 | 29.17 | 35.00 | 24.72 | 38.89 | - | 33.93 |
| **Average** | 45.58 | 34.54 | 47.21 | 33.04 | 41.37 | 31.58 | 47.12 | 43.67 | - |

TABLE IX
TRANSFER ATTACK SUCCESS RATE (TASR) OF THE PROPOSED METHOD ON NWPU DATASET

| Surrogate Model | Target Model | | | | | | | | |
| | AlexNet | VGG16 | ResNet18 | ResNet101 | DenseNet121 | DenseNet201 | Inception-v3 | RegNetX-400MF | **Average** |
|---|---|---|---|---|---|---|---|---|---|
| AlexNet | - | 50.10 | 49.69 | 33.88 | 40.04 | 35.52 | 46.41 | 42.09 | 42.53 |
| VGG16 | 78.95 | - | 58.53 | 48.00 | 53.89 | 52.84 | 60.84 | 54.95 | 58.29 |
| ResNet18 | 80.34 | 63.21 | - | 47.78 | 54.97 | 46.93 | 59.83 | 55.39 | 58.35 |
| ResNet101 | 80.53 | 65.43 | 71.77 | - | 62.58 | **58.86** | 66.08 | 61.49 | 66.68 |
| DenseNet121 | 79.04 | 69.81 | 67.30 | 51.15 | - | 55.14 | 60.17 | 58.91 | 63.07 |
| DenseNet201 | **82.79** | **75.82** | **73.42** | **60.78** | **65.80** | - | **66.23** | **66.67** | **70.22** |
| Inception-v3 | 77.23 | 64.60 | 62.32 | 46.58 | 54.87 | 49.69 | - | 55.07 | 58.62 |
| RegNetX-400MF | 76.36 | 59.83 | 60.04 | 45.19 | 52.72 | 49.16 | 56.49 | - | 57.11 |
| **Average** | 79.32 | 64.11 | 63.30 | 47.62 | 54.98 | 49.73 | 59.44 | 56.37 | - |

### E. Transfer Attack

We conduct transfer attack experiments to evaluate the transferability of adversarial examples generated by our method. During implementation, we first select a neural network model as the surrogate model. Then the adversarial examples generated on this model are used to attack other unknown target models. The resulting TASRs of our method for different target models on UCM and NWPU datasets are shown in Tables VIII-IX.

From Table VIII, it can be observed that using ResNet101 as the surrogate model for transfer attacks yields an average TASR of 47.86%, outperforming all other models. Target models such as AlexNet, ResNet18, and Inception-v3 are more vulnerable to transfer attacks, with average TASRs exceeding 45%. This may be due to the relatively simple structures of these models, which make them less robust against transferred adversarial examples. When comparing the TASRs between pairs of structurally similar but differently layered models, such as ResNet18 with ResNet101 and DenseNet121 with DenseNet201, the higher-layer models, ResNet101 and DenseNet201, achieve TASRs of 55.69% and 44.92% against ResNet18 and DenseNet121. These rates are significantly higher than the reverse transfer attack results (29.63% and 35.96%). Therefore, adversarial examples generated using more complex network structures as surrogate models may have stronger generalization abilities.

In Table IX, the TASRs for various neural networks on the NWPU dataset are generally higher than those on the UCM dataset. This might be due to the larger number of categories in the NWPU dataset, making it more difficult to distinguish different classes, thus enhancing the transferability of adversarial examples on the NWPU dataset. Among different target models, AlexNet has the highest average TASR at 79.32%, indicating it is the most susceptible to transfer attacks.

This may be attributed to its original classification accuracy on the NWPU dataset being the lowest at 86.69%.

### F. Defense Against Perlin Noise Clouds

To further assess the robustness of the proposed method in adversarial defense scenarios, we chose the best-performing black-box method, Square Attack, and the white-box method, BIM, for comparison. First, we use the aforementioned three adversarial attack methods to generate adversarial examples and obtain the original ASRs. Then, the adversarial defense methods Total Variance Minimization (TVM) [56] and Jpeg Compression (JC) [57] are employed to process these adversarial examples, reversing the effects of perturbations. The ASRs after the adversarial defense processing are then reassessed. The parameters for the adversarial defense methods are set as follows: the Bernoulli distribution probability for TVM is set to 0.3, with the norm and lambda parameters set to 2 and 0.5. The L-BFGS-B algorithm is used for optimization, with a maximum of 10 iterations. The image quality level for JC is 50. The experimental results are shown in Fig. 16.

The robustness of the proposed method is slightly better than that of Square Attack. The proposed method demonstrates higher ASRs on AlexNet and Inception-v3 under adversarial defense scenarios. Although the original ASRs of our method on ResNet101, DenseNet121, and DenseNet201 are more than 10% lower than those of Square Attack, the results after adversarial defense processing are similar. This indicates that a lower proportion of adversarial examples generated by our method are successfully defended against. For other models, the two methods perform similarly.

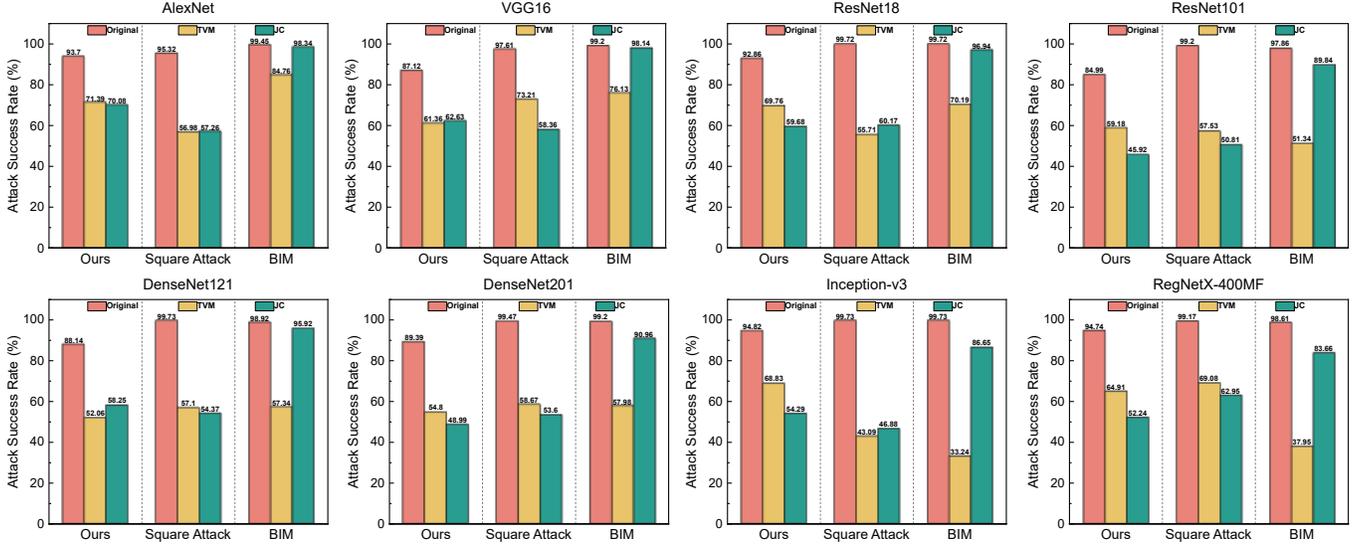

Fig. 16. ASRs of different attack methods against various defense methods on the UCM dataset. In each figure, the red, yellow, and green bars represent the original ASR, the ASR under the TVM defense method, and the ASR under the JC defense method. Each figure shows, from left to right, the ASRs of the proposed method, Square Attack, and BIM.

Comparing the effects of TVM and JC on the various attack methods, it is found that TVM effectively defends against all three attack methods, while JC has very limited defense effectiveness against BIM. This difference can be attributed to two factors. First, TVM randomly selects a set of pixels on the image based on Bernoulli sampling and reconstructs the selected pixels according to the principle of total variance minimization, resulting in an image with minimal perturbations. This is a randomized defense, whereas JC removes minor pixel value changes through data quantization, a deterministic denoising procedure, making TVM generally more effective in defense. Second, the proposed method and Square Attack generate specific patterns of perturbations, such as clouds or blocks, unlike BIM, which directly applies adversarial perturbation. Thus, they may lose more perturbation effect when facing JC.

## IV. CONCLUSION

In this paper, we propose a novel black-box attack method to generate cloud adversarial examples for remote sensing image classification based on Perlin noise. To simplify the generation of the Perlin noise cloud mask into a black-box optimization problem, we design PGGN to obtain grids of different sizes with gradient vectors and use the DE algorithm to achieve a query-based black-box attack. Compared to existing methods, our approach leverages the atmospheric characteristics of remote sensing images by simulating clouds to generate adversarial examples. Such perturbations are more rational and align with human cognition. Extensive experiments on two remote sensing image datasets demonstrate the effective attack capability and query efficiency of the proposed method across different neural networks. Furthermore, we discuss the attack selectivity based on the distribution of adversarial example categories and explain the potential impact of the generated clouds on neural networks using CAMs. Additionally, we evaluate the transfer attack capability of the generated adversarial examples and their robustness in adversarial defense scenarios.

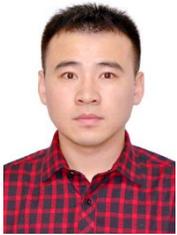
**Fei Ma** (Member, IEEE) received the B.S., M.S., and Ph.D. degrees in electronic and information engineering from the Beijing University of Aeronautics and Astronautics (BUAA), Beijing, China, in 2013, 2016, and 2020 respectively. He is currently working with the College of Information Science and Technology, Beijing University of Chemical Technology, Beijing, China, as an Associate Professor. His research interests include radar signal processing, image processing, machine learning, and target detection.

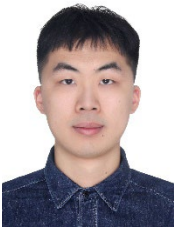
**Yuqiang Feng** received the B.E. degree in artificial intelligence from the Beijing University of Chemical Technology, Beijing, China, in 2024. His research interests include remote sensing image interpretation, machine learning, and adversarial attack.

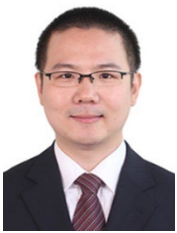
**Fan Zhang** (Senior Member, IEEE) received the B.E. degree in communication engineering from the Civil Aviation University of China, Tianjin, China, in 2002, the M.S. degree in signal and information processing from Beihang University, Beijing, China, in 2005, and the Ph.D. degree in signal and information processing from Institute of Electronics, Chinese Academy of Sciences, Beijing, in 2008. He is a Full Professor of electronic and information engineering at the Beijing University of Chemical Technology, Beijing. His research interests are remote sensing image processing, high-performance computing, and artificial intelligence.

Dr. Zhang is also an Associate Editor of IEEE Access and a Reviewer of the IEEE Transactions on Geoscience and Remote Sensing, the IEEE Journal of Selected Topics in Applied Earth Observations and Remote Sensing, the IEEE Geoscience and Remote Sensing Letters, and the Journal of Radars.

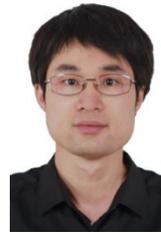
**Yongsheng Zhou** (Member, IEEE) received the B.E. degree in communication engineering from Beijing Information Science and Technology University, Beijing, China, in 2005, and the Ph.D. degree in signal and information processing from the Institute of Electronics, Chinese Academy of Sciences, Beijing, China, in 2010. He was with the Academy of Opto-Electronics, Chinese Academy of Sciences between 2010 and 2019, and is currently a Professor of Electronic and Information Engineering with the College of Information Science and Technology, Beijing University of Chemical Technology, Beijing, China. His general research interests include target detection and recognition from microwave remotely sensed images, digital signal and image processing.